\documentclass[10pt,twocolumn]{article}

\usepackage[
  a4paper,
  top=0.75in,
  bottom=0.8in,
  left=0.7in,
  right=0.7in
]{geometry}
\usepackage[T1]{fontenc}
\usepackage[utf8]{inputenc}
\usepackage{microtype}

\usepackage{amsmath}
\usepackage{amssymb}
\usepackage{amsthm}
\newtheorem{definition}{Definition}

\usepackage{graphicx}
\usepackage{booktabs}
\usepackage{multirow}
\usepackage{tabularx}
\usepackage{placeins}
\usepackage{capt-of}
\renewcommand{\arraystretch}{1.1}
\usepackage{listings}
\usepackage{xcolor}
\definecolor{codegreen}{rgb}{0,0.6,0}
\definecolor{codegray}{rgb}{0.5,0.5,0.5}
\definecolor{codepurple}{rgb}{0.58,0,0.82}
\definecolor{backcolour}{rgb}{0.95,0.95,0.92}

\usepackage[numbers,sort&compress]{natbib}
\usepackage[hidelinks]{hyperref}

\title{TS-Mob: Social and Geographical-Aware Time Series Foundation-Model Framework for Human Mobility Prediction}

\author{%
Massimiliano Luca$^{1}$ \quad
Ciro Beneduce$^{1,2}$ \quad
Bruno Lepri$^{1}$\\[0.6em]
\small $^{1}$MobS Lab, Bruno Kessler Foundation, Trento, Italy\\
\small $^{2}$Department of Computer Science, University of Trento, Trento, Italy\\[0.3em]
\small \texttt{mluca@fbk.eu, cbeneduce@fbk.eu, lepri@fbk.eu}
}
\date{}

\begin{document}
\maketitle

\begin{abstract}
Short-term forecasting of aggregated human mobility flows supports urban planning, intelligent transportation, and emergency response, yet existing models often require substantial mobility history and learn spatial structure only implicitly through grids or graphs. Time series foundation models provide strong temporal priors but lack explicit geographic and social conditioning for origin-destination interactions. We introduce TS-Mob, a framework that conditions a fine-tuned time series foundation model (TimesFM) on a gravity-inspired destination-attractiveness signal computed entirely from open data (living population, centroid distances, and Overture POI counts) together with weather covariates. This couples the temporal prior of large-scale pre-training with an explicit, interpretable representation of the structural drivers of mobility, without training a fully spatiotemporal model from scratch. We evaluate TS-Mob on three benchmarks spanning different spatial scales: Bike New York City, Taxi Beijing, and a nation-scale Spain origin-destination matrix estimated through mobile phone data. Across all metrics (RMSE, MAE, CPC), TS-Mob consistently outperforms classical, deep spatio-temporal, and foundation-model baselines in both zero-shot and fine-tuned settings. Gains over the strongest classical baseline reach 67.04\% lower MAE and 91.66\% higher CPC, and TS-Mob remains the best method even against the strongest fine-tuned foundation baseline, with improvements up to 3.43\% lower RMSE and 4.69\% higher CPC. Stratified analyses further show robustness across different temporal regimes, like weekdays/weekends and multi-horizon forecasting.
\end{abstract}

\section{Introduction}
\label{sec:intro}

Understanding and forecasting human mobility has major implications for public safety, migration analysis, on-demand services, pollution monitoring, epidemic diffusion, and traffic optimization \cite{luca2021survey, bohorquez2009common, colizza2007modeling, bettencourt2007growth, alessandretti2022human}. Crowd flow prediction aims to predict, for each origin, the number of individuals moving from/to every possible destination in the next time interval. This capability supports operational planning and rapid response, from fleet rebalancing and traffic management to anticipating demand spikes and detecting mobility anomalies \cite{luca2021survey, barbosa2018human}.

Accurate crowd flow forecasting is challenging because mobility patterns arise from the interaction of time-varying dynamics and stable structural constraints \cite{luca2021survey,zhang2017deep}. Temporal signals include strong daily and weekly regularities, seasonal trends, and exogenous drivers such as weather conditions \cite{zhang2017deep,yao2018deep,liu2018attentive,luca2021survey,barbosa2018human}; structural geographic and social signals capture aspects like distance decay, population mass, and destination opportunities that are known shape where people go \cite{barbosa2018human, simini2021deep, zipf1946p, simini2012universal}. Existing approaches often entangle these aspects. Classical statistical models \cite{hamilton2020time} are efficient and provide solid baselines, but tend to saturate when non-linear spatio-temporal interactions are required \cite{luca2021survey}. Deep spatio-temporal models, including convolutional architectures on regular grids \cite{zhang2017deep, yao2018deep, liu2018attentive} and graph-based approaches over regions \cite{cardia2022enhancing, ul2025crowd, lan2022dstagnn, jiang2022graph}, can learn complex dependencies and achieve strong performance, yet they frequently require substantial mobility history and rely on dataset-specific geographic representations  \cite{zhang2017deep, yao2018deep, lan2022dstagnn, cardia2022enhancing, ul2025crowd, lan2022dstagnn, jiang2022graph, luca2021survey}. These requirements limit portability across cities, spatial partitions, and data-scarce contexts.
Moreover, approaches based on various types of foundation models are revolutionising both the panorama of predictive models for individual \cite{gong2024mobility, wang2023would, beneduce2025large, feng2025agentmove, chen2025enhancing, xue2023promptcast, huang2025vision}, and collective \cite{woo2024unified, das2024decoder, jin2023time, xue2023promptcast} level mobility patterns.
More recently, learning-based approaches have expanded the modelling landscape for both individual mobility prediction~\cite{gong2024mobility, wang2023would, beneduce2025large, feng2025agentmove, chen2025enhancing, xue2023promptcast, huang2025vision} and collective mobility forecasting~\cite{woo2024unified, das2024decoder, jin2023time, xue2023promptcast}. In parallel, recent work on mobility networked time series and benchmark datasets has emphasised that mobility forecasting requires accounting for both temporal evolution and interaction structure across spatial units~\cite{nam2025bi, na2025mobility}.
Among these approaches, time series foundation models \cite{woo2024unified, das2024decoder} or hybrid text-based approaches \cite{jin2023time, xue2023promptcast} provide a complementary opportunity: pre-trained on large-scale heterogeneous time series, they offer strong temporal priors that can be adapted to new domains with relatively limited supervision \cite{woo2024unified, das2024decoder}. 
However, generic time series foundation models are not designed to incorporate the explicit geographic and social structure that governs crowd flow interactions, and therefore may overlook key drivers such as distance decay and destination attractiveness. This paper introduces TS-Mob, a framework that couples time series foundation models with an explicit and interpretable representation of geographic and social structure. The key idea is to condition a temporal forecaster on a gravity-inspired destination-attractiveness signal. TS-Mob includes a geographic- and social-aware block that estimates, for each origin-destination pair, an attractiveness value using the living population, the centroid-to-centroid Haversine distance \cite{barbosa2018human}, and Overture \footnote{https://overturemaps.org/} POI counts. These OD-specific attractiveness values are then provided as covariates for the fine-tuning of a temporal block based on TimesFM \cite{das2024decoder}, together with weather features (i.e., millimetres of precipitation and temperature expressed in degrees Celsius). This design injects stable structural knowledge into the temporal model without requiring a fully spatiotemporal architecture to be trained from scratch, and it produces an interpretable signal that can be inspected to understand destination preference patterns induced by population, activities, and separation.

We evaluate TS-Mob on three real-world mobility datasets spanning different spatial scales and temporal resolutions: Bike New York City \cite{zhang2017deep}, Taxi Beijing \cite{zhang2017deep}, and nation-scale Spain OD \cite{MITMA2023, beneduce2025pyspainmobilitypythonpackageaccess}. Across all datasets and evaluation metrics, namely Root Mean Square Error (RMSE), Mean Absolute Error (MAE), and Common parts of Commuters (CPC) \cite{lenormand2016systematic, luca2021survey, barbosa2018human}, TS-Mob consistently outperforms classical baselines, widely adopted deep spatio-temporal models, and strong foundation-model baselines. Improvements are particularly pronounced over classical methods (up to 67.04\% lower MAE and 91.66\% higher CPC), and remain consistent even relative to the strongest fine-tuned foundation baseline (up to about 3.43\% lower RMSE and 4.69\% higher CPC). Gains are especially pronounced in CPC, which measures recovery of the full OD flow structure rather than pointwise error reduction, and which is the metric most directly relevant to operational mobility applications \cite{lenormand2016systematic, luca2021survey, barbosa2018human}. Additional stratified analyses show that TS-Mob remains robust under temporal regime shifts, including weekdays versus weekends and multi-horizon forecasting. Our contributions are the following: 
\begin{itemize}
    \item We propose TS-Mob, a mobility forecasting framework that conditions a time series foundation model on an interpretable gravity-inspired attractiveness signal without proprietary or dataset-specific representations. 
    \item We benchmark TS-Mob against classical, deep spatio-temporal, and fine-tuned foundation-model baselines and consistently outperform them across all datasets and metrics, with gains reflected in both pointwise error and OD structural recovery (CPC). 
\end{itemize}

To support reproducibility, we release the code and scripts used to run our experiments at {\color{blue}\url{https://anonymous.4open.science/r/tsmob-715C/}}. The remainder of the paper is organized as follows. Section~\ref{sec:problem_definition} formalizes the problem.  Section~\ref{sec:tsmob} presents TS-Mob. In Section~\ref{sec:methodology}, we introduce the selected baselines, evaluation metrics, and datasets. Section~\ref{sec:results} reports the main experimental results, followed by a temporal regime analyses in Section~\ref{sec:temporal_pattern_analysis} and ablation study in \ref{sec:ablation}. Finally, we conclude and discuss future directions in Section \ref{sec:conclusion}.

\section{Problem Definition}
\label{sec:problem_definition}

Human mobility modeling often relies on dividing a geographical area into discrete regions and analyzing aggregate movement patterns between them. This section introduces the formal definitions necessary for crowd flow prediction and presents our novel formulation of the problem as a univariate time series forecasting task.

\subsection{Spatial Tessellation}

Let $R$ be a geographical area and $G$ a set of polygons. $G$ is called a \textit{tessellation} if the following properties hold:
\begin{enumerate}
    \item $G$ contains a finite number of polygons (i.e., tiles) $l_i$, so that $G = \{l_i: i=1, ..., n\}$;
    \item Locations are not overlapped, that is $l_i \cap l_j = \emptyset, \forall i \neq j$;
    \item The union of all the locations entirely covers $R$, i.e. $\bigcup_{i=1}^{n}l_i = R$.
\end{enumerate}
Tessellations allow us to map data points into a finite number of tiles within the area, instead of working with raw positional coordinates. Tiles can be regular geometric shapes (such as squares, triangles, or hexagons) or irregular ones (such as census tracts or administrative units). Properties (2) and (3) ensure that each point in the region is assigned to only one tile.

\subsection{Crowd Flow}

Mobility flows represent aggregated movements among geographic locations, typically modeled using an Origin-Destination (OD) tensor. Let $T \in \mathbb{N}^{t \times n \times n}$ denote the OD tensor, where $t$ is the number of time intervals, and $n$ is the number of tiles in the tessellation. The entry $T_{\tau, i, j}$ corresponds to the number of individuals moving from origin tile $i$ to destination tile $j$ during time interval $\tau$.

To characterize the dynamics within each tile, we define the crowd flow as the aggregate number of individuals entering or leaving each tile during a given time interval. Given a set of individual trajectories, the incoming and outgoing crowd flows between tiles $(i,j)$ during time interval $t$ are given by:
$$
    in_t^{(i,j)} = \sum_{K\in Q} |\{t > 1 : (i,j) \notin q_{K}^{t-1} \wedge (i,j) \in q_{K}^{t}\}| $$ $$
    out_t^{(i,j)} = \sum_{K\in Q} |\{t > 1 : (i,j) \in q_{K}^{t} \wedge (i,j) \notin q_{K}^{t+1}\}|
$$
where $Q$ denotes the set of all individual trajectories, and $q_{T}^{t}$ represents the set of tiles traversed by trajectory $K$ at time $t$.

\subsection{Crowd Flow Prediction}

Traditionally, the crowd flow prediction problem is formulated as follows:

\begin{definition}[Crowd Flow Prediction]
\label{def:crowd_flow_prediction_problem}
Given a spatial tessellation $R$ composed of $n$ tiles, and the crowd flows for each tile over $t$ time intervals, crowd flow prediction consists in forecasting $X_{t+c}$, where $c \in \mathbb{N}$, given the historical crowd flows $\{X_{i} : i = 1, ..., t\}$.
\end{definition}

In this work, we propose a novel, slightly different, formulation of the crowd flow prediction problem. Instead of learning the joint dynamics of the entire OD tensor, we decompose the problem into a set of independent univariate time series forecasting tasks--one for each origin tile.

Given a spatial tessellation $R$ composed of $n$ tiles, let $T \in \mathbb{N}^{t \times n \times n}$ be the historical crowd flows, where $T_{\tau, i, j}$ denotes the number of individuals moving from origin tile $i$ to destination tile $j$ at time interval $\tau$. For each origin tile $i$, define the sequence of outgoing flows over time as $S^{(i)} = \{ \mathbf{s}_1^{(i)}, \dots, \mathbf{s}_t^{(i)} \}$, with $\mathbf{s}_\tau^{(i)} = [T_{\tau, i, 1}, ..., T_{\tau, i, n}]$.
    
The per-origin time series crowd flow prediction problem consists in, for each origin $i$, forecasting the next vector of outgoing flows $\mathbf{s}_{t+1}^{(i)}$ given the historical series $\{ \mathbf{s}_1^{(i)}, ..., \mathbf{s}_t^{(i)} \}$.

This reformulation transforms the spatiotemporal crowd flow prediction task into $n(n-1)$ independent univariate forecasting problems, whose predictions can be regrouped by origin $i$ to recover the full outgoing-flow vector $\mathbf{s}_{t+1}^{(i)}$.

\section{TS-Mob: Temporal Forecasting with Geographic- and Social-Aware Attractiveness Conditioning}

\label{sec:tsmob}
\begin{figure*}[ht!]
    \centering
    \includegraphics[width=0.9\linewidth]{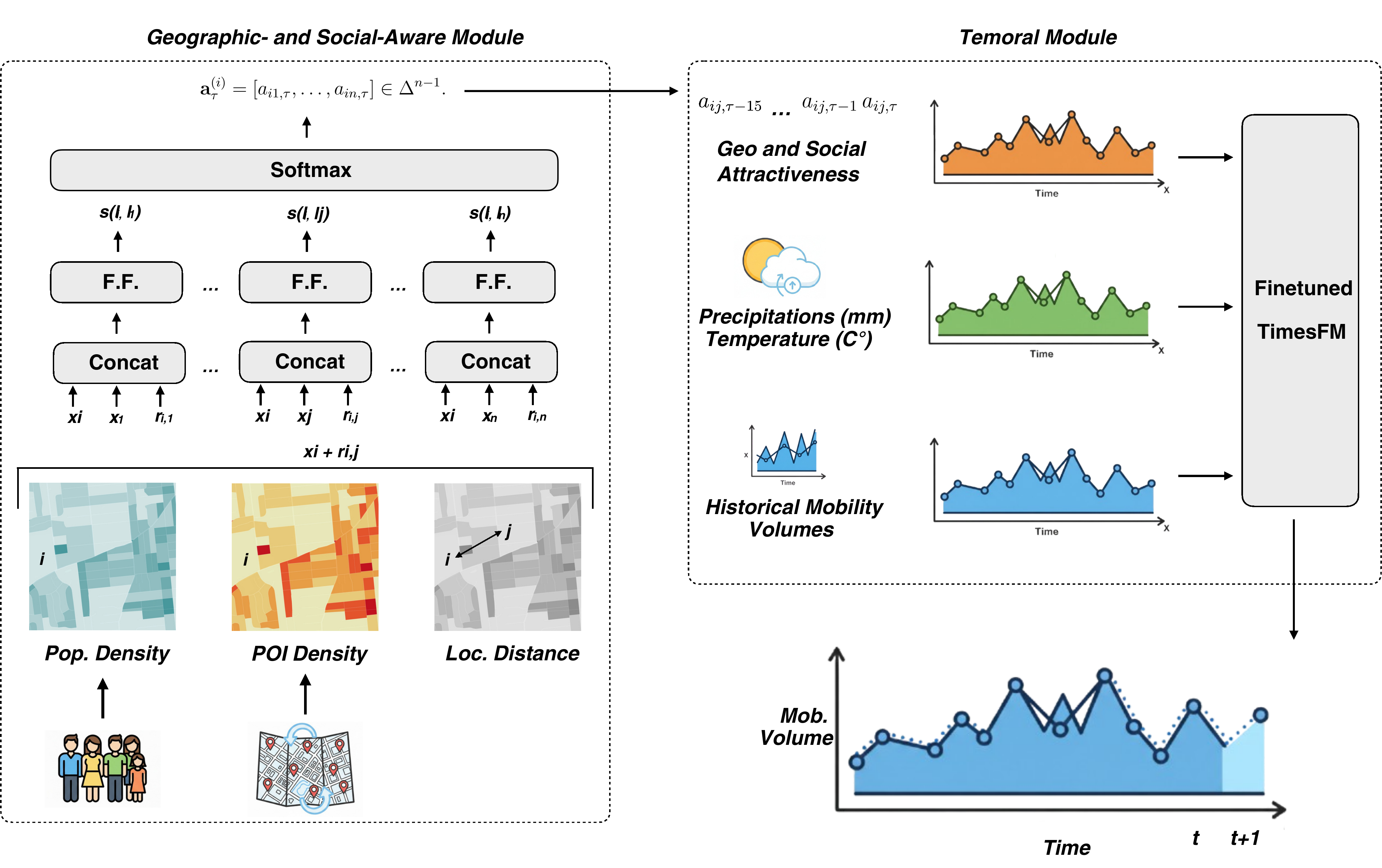}
    \caption{TS-Mob pipeline. A gravity-inspired geographic/social block computes a time-varying destination-attractiveness signal $a_{ij,\tau}$ from open data (living population, Overture POIs, and distance). A TimesFM-based temporal forecaster then predicts the next OD flow using historical mobility volumes together with $a_{ij,\tau}$ and exogenous covariates (e.g., weather).}

    \label{fig:framework}
\end{figure*}
TS-Mob combines two complementary components: (i) a \emph{geographic- and social-aware block} that estimates, for each origin, an interpretable destination-attractiveness profile from open geographic and social signals; and (ii) a \emph{temporal block} based on a fine-tuned Time Series Foundation Model (TimesFM \cite{das2024decoder}) that predicts future OD flows while being explicitly conditioned on such attractiveness information. The proposed framework is also summarized in Figure \ref{fig:framework}. This design injects structural geographic/social knowledge into the temporal forecaster as an origin-specific covariate (one entry per possible destination), enabling geographically-aware predictions without training a fully spatiotemporal model from scratch.

\subsection{Notation and Problem Setup}

Let $T \in \mathbb{N}^{t \times n \times n}$ be the OD tensor, where $T_{\tau,i,j}$ denotes the number of individuals moving from origin tile $i$ to destination tile $j$ at time interval $\tau$. 


For each OD pair $(i,j)$ with $i \neq j$, define the univariate time series
$$
y^{(i,j)} = \{ T_{1,i,j},\ T_{2,i,j},\ \dots,\ T_{t,i,j} \}.
$$
Let $\mathbf{x}_\tau \in \mathbb{R}^{d_x}$ denote time-varying covariates observed at time $\tau$ (e.g., weather). We forecast $c \ge 1$ steps ahead using a lookback window of length $L$.

For each OD pair $(i,j)$, let $\mathbf{u}_{ij}\in\mathbb{R}^{d_u}$ be static geographic/social attributes computed from open data (e.g., populations, distances, POI-based proxies of attractiveness).

\subsection{Geographic- and Social-Aware Block: Gravity-Inspired Destination Attractiveness}
\label{sec:tsmob_spatial_block}

The geographic- and social-aware block provides the temporal forecaster with an origin-specific, destination-level covariate that quantifies how \emph{structurally attractive} each candidate destination is, based on open geographic and activity signals. We start from the classic gravity-model view of destination choice, where selecting a destination among $n$ alternatives can be expressed as a multinomial logistic regression. This formulation is formally equivalent to a linear neural network followed by a softmax layer; we extend it by introducing non-linearities and additional hidden layers, obtaining a deep gravity-inspired parametrization of the problem as suggested in \cite{simini2021deep}.

For each location $l_i$, we define a location feature vector $\mathbf{x}_i$ built entirely from open data. In our implementation, $\mathbf{x}_i$ includes (i) the \emph{living population} of the tile (i.e., the actual number of people present in the area, varying over time rather than the registered resident population); and (ii) the number of Points of Interest (POIs) extracted from Overture Maps within the tile polygon, disaggregated into 16 top-level categories (listed in Appendix~\ref{app:overture_categories}). For each origin-destination pair $(i,j)$ we also compute a separation term $r_{ij}$ as the Haversine distance between the centroids of the two tiles.

Given an origin tile $l_i$, the model considers all $n$ possible destinations $\{l_j\}_{j=1}^n$ and constructs one input vector per candidate destination by concatenating origin features, destination features, and distance:
$$
\mathbf{x}(l_i,l_j) \;=\; \mathrm{concat}\!\big[\mathbf{x}_i,\ \mathbf{x}_j,\ r_{ij}\big],
\qquad j=1,\dots,n.
$$
Concretely, each $\mathbf{x}(l_i,l_j)$ contains: population at the origin, population at the destination, the centroid-to-centroid distance, the 16 POI-count features of the origin tile, and the 16 POI-count features of the destination tile.

All candidate vectors $\mathbf{x}(l_i,l_j)$ are processed in parallel by the same feed-forward neural network, with parameters shared across OD pairs. The attractiveness module is trained as a destination-choice model, where the network learns to assign a probability distribution over candidate destinations for each origin, following the deep gravity formulation in~\cite{simini2021deep}.

To avoid temporal leakage, the attractiveness module is trained only using OD flows from the temporal training window. Let $\mathcal{T}_{\mathrm{train}}$ denote the chronological training period. We aggregate training-period flows as
\[
y(l_i,l_j)=\sum_{\tau\in \mathcal{T}_{\mathrm{train}}} T_{\tau,i,j},
\qquad
O_i=\sum_{j\neq i} y(l_i,l_j).
\]
The target destination distribution for origin $i$ is therefore $y(l_i,l_j)/O_i$. The random 50/50 origin split is applied only within $\mathcal{T}_{\mathrm{train}}$: 50\% of origins are used to train the attractiveness module and the remaining 50\% are used for model selection. The held-out future forecasting test period is never used to fit or validate the attractiveness module.

The model is optimized with the cross-entropy loss
\[
H=-\sum_i\sum_{j\neq i}\frac{y(l_i,l_j)}{O_i}\log p_{ij},
\]
using RMSprop with momentum $0.9$ and learning rate $5\cdot 10^{-6}$. We use a seven-layer fully connected network with LeakyReLU nonlinearities to map each candidate vector to a scalar score $s(l_i,l_j)\in(-\infty,+\infty)$:
$$
\mathbf{z}^{(0)}(l_i,l_j) = a\!\left(\mathbf{W}^{(0)}\mathbf{x}(l_i,l_j)+\mathbf{b}^{(0)}\right), \nonumber
$$
$$
\mathbf{z}^{(h)}(l_i,l_j) = a\!\left(\mathbf{W}^{(h)}\mathbf{z}^{(h-1)}(l_i,l_j)+\mathbf{b}^{(h)}\right), \quad h=1,\dots,5, \nonumber
$$
$$
s(l_i,l_j) = \mathbf{w}^\top \mathbf{z}^{(6)}(l_i,l_j) + b. 
$$
Scores are then converted into an origin-specific attractiveness profile over the $n$ candidate destinations using a softmax normalization:
$$
a_{ij,\tau}
\;=\;
\frac{\exp\big(s(l_i,l_j)\big)}{\sum_{k=1}^{n}\exp\big(s(l_i,l_k)\big)},
\qquad
\mathbf{a}^{(i)}_{\tau}=[a_{i1,\tau},\dots,a_{in,\tau}] \in \Delta^{n-1}.
$$
Intuitively, $a_{ij, \tau}$ represents the attractiveness potential of destination $j$ from origin $i$, as induced by population mass measured at time $\tau$, POI-based activity opportunities, and distance decay.
The resulting vector $\mathbf{a}^{(i)}_{\tau}$ is a geography- and social-aware destination-attractiveness covariate: it assigns higher values to destinations that are structurally plausible given the origin context, the destination ``mass'' and activity proxies, and the separation between tiles. In TS-Mob, $\mathbf{a}^{(i)}_{\tau}$ is used as an additional covariate for the temporal block, allowing the fine-tuned TimesFM to model temporal dynamics while being explicitly conditioned on an interpretable representation of geographic and social attractiveness.

\subsection{Temporal Block: TimesFM Conditioned on Attractiveness Covariates}
\label{sec:tsmob_temporal_block}

The temporal block is responsible for learning the time-evolving dynamics of crowd flows while being explicitly conditioned on static geographic and social structure. For each origin-destination pair $(i,j)$, we model the mobility flow as a univariate time series $\{T_{\tau,i,j}\}_{\tau}$ and forecast future values using a fine-tuned version of TimesFM. Crucially, the temporal forecaster receives, alongside the historical flow values, a set of covariates that include (i) an OD-specific attractiveness signal produced by the gravity-inspired block and (ii) exogenous temporal signals such as weather conditions (i.e., millimeters of precipitation and temperature in degrees Celsius) and calendar effects (i.e., weekend or weekday marker).

Given an OD pair $(i,j)$ and a reference time $t$, we build an input window of fixed length $L=32$ by collecting the historical flow values and covariates from $t-L$ to $t$. Let $y_{\tau}^{(i,j)} = T_{\tau,i,j}$ denote the observed flow count between $i$ and $j$ at time $\tau$. Let $a_{ij}$ denote the attractiveness of destination $j$ given origin $i$ computed by the geographic- and social-aware block (Section~\ref{sec:tsmob_spatial_block}). $a_{ij, \tau}$ provides to the temporal module an OD-specific covariate stream aligned with the temporal input window. Finally, we include time-varying exogenous covariates $\mathbf{w}_{\tau}$, which in our case comprise millimeters of precipitation, perceived temperature in Celsius, and the hour of the day (encoded as a temporal feature). We denote the resulting covariate vector at time $\tau$ as
$$
\mathbf{c}^{(i,j)}_{\tau} \;=\; \big[a_{ij,\tau},\ \texttt{precip}_{\tau},\ \texttt{temp}_{\tau},\ \texttt{hour}_{\tau}\big].
$$

The temporal model receives the context window
$$
\mathcal{D}^{(i,j)}_{t} \;=\; \left\{\left(y_{\tau}^{(i,j)},\ \mathbf{c}^{(i,j)}_{\tau}\right)\right\}_{\tau=t-L}^{t},
$$
and outputs a forecast for the next time steps. The aforementioned data are provided during fine-tuning and during inference time. In the standard crowd-flow prediction setting considered in this work, the primary target is the one-step-ahead flow:
$$
\widehat{y}^{(i,j)}_{t+1}
\;=\;
f_{\theta}\!\left(\mathcal{D}^{(i,j)}_{t}\right),
$$
where $f_{\theta}$ is TimesFM after fine-tuning. Further details on fine-tuning settings are available in Appendix \ref{app:times_finetuned} 

Time is discretized according to the native aggregation of each mobility dataset: hourly for the Spain and Bike New York City dataset, 30-minute bins for Taxi Beijing. The covariate streams are aligned to the same temporal granularity as the mobility counts. We train the temporal forecaster to predict raw flow counts directly, without applying variance-stabilizing transformations, and evaluate performance on the original count scale.

The temporal block is fine-tuned with a mean squared error objective. For a horizon $H=1$ the training loss is
$$
\mathcal{L}_{\text{temp}}(\theta)
\;=\;
\sum_{h=1}^{H}
\left(
y^{(i,j)}_{t+h}
-
\widehat{y}^{(i,j)}_{t+h}
\right)^2.
$$

Notably, TimesFM has been fine-tuned with SGD \cite{ruder2016overview} with linear warmup of 25 epochs, and cosine decay with a learning rate of 5e-4 with a momentum of 0.9.

Finally, we follow standard practice in mobility modeling by excluding self-flows (i.e., $i=j$) from both training and evaluation \cite{luca2021survey}.
Operationally, the gravity-inspired block is trained first to compute attractiveness values $a_{ij}$, which are then treated as fixed covariates when fine-tuning TimesFM. At inference time, the same attractiveness values and covariate streams are provided to the temporal block together with the most recent $L=32$ observations to generate geographic- and social-aware forecasts.

\subsection{Inference and Interpretability}
\label{sec:tsmob_inference_interpretability}

At inference time, TS-Mob operates in two steps. First, the geographic- and social-aware block is used to compute an OD-specific attractiveness value $a_{ij}$ for every admissible origin-destination pair $(i,j)$ (with $i\neq j$), using only open static inputs (living population, Overture POI counts at origin and destination, and centroid-to-centroid Haversine distance). These attractiveness values are treated as covariates for the temporal forecaster.

Second, for each OD pair $(i,j)$ and reference time $t$, the temporal block receives a context window of length $L=32$ consisting of the historical flow counts $\{T_{\tau,i,j}\}_{\tau=t-L}^{t}$ and aligned covariate streams. In particular, the covariates include the OD-specific attractiveness $a_{ij}$ (provided as a scalar covariate stream over the window), precipitation (mm), perceived temperature (Celsius), and calendar time features (hour of day). The model then outputs the one-step-ahead prediction $\widehat{T}_{t+1,i,j}$, which corresponds to the expected number of individuals moving from origin $i$ to destination $j$ at the next time interval.

In addition to improving predictive accuracy, TS-Mob offers an interpretable handle on the geographic and social structure of mobility through the learned attractiveness scores. Because $a_{ij}$ is computed from explicit, human-readable factors (population mass, distance decay, and POI-derived activity opportunities), it can be inspected directly to characterize which destinations are structurally most attractive from each origin, and to compare attractiveness patterns across different origins (e.g., central vs. peripheral areas). Moreover, by analyzing how attractiveness varies as a function of distance and destination activity/population proxies, the framework enables a principled assessment of the learned trade-off between separation costs and destination ``mass'', providing a geography- and social-aware layer of explanation for the temporal predictions.

\section{Methodology}
\label{sec:methodology}

This section describes the experimental methodology adopted to evaluate TS-Mob and the competing approaches. We first outline the baseline methods considered in our study, then describe the mobility datasets and the evaluation metrics used throughout the paper. 
For reproducibility, Appendix~\ref{app:repro_details} reports data preprocessing, baseline training details, training of the geographic and social module, TimesFM fine-tuning, and computational settings.

\subsection{Baselines}

We compare TS-Mob against three groups of baselines: traditional methods, deep learning models, and foundation-model baselines.

\paragraph{Traditional methods.}
We include \textbf{Moving Average (MA)}, \textbf{Auto-Regressive Integrated
Moving Average (ARIMA)}~\cite{hamilton2020time}, and \textbf{Vector
Auto-Regressive (VAR)}~\cite{hamilton2020time} as classical temporal
forecasting baselines. These methods provide non-neural references for
forecasting each OD flow from its historical observations, without using
learned spatial representations or pretrained foundation models.

\paragraph{Deep learning models.}
This group includes sequential neural forecasters and spatio-temporal architectures. We include \textbf{RNN} \cite{hopfield1982neural}, \textbf{LSTM} \cite{hochreiter1997long}, and \textbf{GRU} \cite{cho2014learning} as neural baselines aligned with the OD-pair time-series formulation. We also include widely used spatio-temporal crowd-flow models: \textbf{ST-ResNet}~\cite{zhang2017deep}, \textbf{DMVSTNet}~\cite{yao2018deep}, \textbf{ACMF}~\cite{liu2018attentive}, \textbf{CrowdNet}~\cite{cardia2022enhancing}, \textbf{DSTAGNN}~\cite{lan2022dstagnn}, and \textbf{MSAGGN}~\cite{ul2025crowd}. CNN-based grid models are not reported on Spain OD because this dataset is defined over irregular administrative regions rather than a regular grid.

\paragraph{Foundation-model baselines.}
This group includes general-purpose time-series and language-model-based forecasting approaches. We evaluate \textbf{TimesFM}~\cite{das2024decoder}, \textbf{Moirai}~\cite{woo2024unified}, and \textbf{TimeLLM}~\cite{jin2023time}. 
In the main benchmark table, foundation-model baselines are evaluated both in their zero-shot setting and fine-tuned setting to enhance fair comparison. 

Training details and hyperparameter choices for baselines are reported in Appendix~\ref{app:baselines}. In addition, computational settings are described in Appendix \ref{app:Computational_settings}
Lastly, classical spatial interaction models, namely Gravity ~\cite{zipf1946p,barbosa2018human} and Radiation ~\cite{simini2012universal}, are reported separately in Appendix~\ref{app:spatial_interaction_baselines}  as spatial-interaction references.

\subsection{Mobility Datasets}
\label{sec:datasets}

The \textit{Citi Bike System} dataset \footnote{https://ride.citibikenyc.com/system-data} \cite{zhang2017deep} describes trips recorded by the New York bike-sharing system from 2013 to date. We consider trips from April to September 2014 because it is the range of dates usually used in the literature to test crowd flow prediction methods \cite{luca2021survey}. Data are tesselled into a $16,8$ as described in \cite{zhang2017deep}

The \textit{Taxi Beijing} dataset is based on T-Drive \cite{yuan2011t}. It was collected by Microsoft in the area of Beijing, China, and it contains the GPS location of 10,357 taxis sampled every 177 seconds. The data were collected over one week in February 2008. Data are tesselled into a $32,32$  and temporally aggregated into 30-minute blocks as described in \cite{zhang2017deep}.

In addition to the two city-level datasets, we retrieve three months of data (February 1st 2023 - April 30th 2023) at the district level \cite{MITMA2023}. We use the data from the beginning of February to April 15th for training the selected baselines, and we test the performance on the last 15 days. 
Data are sampled hourly and are retrieved and processed using pySpainMobility \cite{beneduce2025pyspainmobilitypythonpackageaccess}. 
Note that, as the data comes with an irregular tessellation, we will evaluate a limited number of models on this dataset. The main reason is that models based on convolutional networks only deal with regular tessellation shapes. 

Data preprocessing steps, including the temporal splitting protocol used across datasets, are reported in Appendix~\ref{app:data_prepro}.

\subsection{Evaluation Metrics}
\label{sec:evaluation_metrics}
The performance of crowd flow predictors is evaluated as the similarity between the predicted heatmap of crowd flows and the real one. 

In our experiments, we adopt Root Mean Squared Error (RMSE), the most used metric to evaluate crowd flow prediction \cite{luca2021survey, barbosa2018human}:
$$\text{RMSE} = \sqrt{\frac{1}{n} \sum_{i=1}^{n} ( T_{ij}-\hat{T}_{ij})^{2}}$$
where $n$ is the number of predictions, $\hat{y}_i$ indicates the predicted value and $y_i$ the actual value.
We exploit the RMSE to evaluate the model performance on both the crowd flow prediction problem and on the flow prediction problem.
We evaluate the goodness of the predictions for the flow prediction problem using the Common Part of Commuters (CPC) \cite{ luca2021survey, barbosa2018human, lenormand2016systematic}:
$$
CPC(\hat{T}, T) = \frac{2\sum_{i,j}min(\hat{T}_{ij}, T_{ij})}{\sum_{i,j}\hat{T_{ij}} + \sum_{ij}T_{ij}}
$$
where $\hat{T}_{ij}$ is the flow from region $i$ to region $j$ predicted by the model and $T_{ij}$ is the actual flow from region $i$ to region $j$.
CPC ranges between 0 and 1: if two adjacency matrices do not have any flows in common, the CPC value is 0. 
CPC is 1 if the sets of flows are identical.
We further assess the predictive accuracy of the models using the Mean Absolute Error (MAE) \cite{luca2021survey, barbosa2018human}, a widely adopted metric for regression tasks. MAE is defined as:
$$
MAE(\hat{T}, T) = \frac{1}{N} \sum_{i,j} \left| \hat{T}{ij} - T{ij} \right|
$$
where $\hat{T}{ij}$ and $T{ij}$ denote the predicted and actual flows from region $i$ to region $j$, respectively, and $N$ is the total number of origin-destination pairs. MAE measures the average magnitude of the errors in the predictions, without considering their direction. Lower values indicate better predictive performance, with an MAE of 0 denoting perfect predictions.

\section{Crowd Flow Forecasting Performances}
\label{sec:results}

\begin{table*}[t]
\centering
\small
\setlength{\tabcolsep}{8pt}
\renewcommand{\arraystretch}{1.09}
\resizebox{0.9\textwidth}{!}{%
\begin{tabular}{l l|ccc|ccc|ccc}
\toprule
\multicolumn{2}{c|}{} 
& \multicolumn{3}{c|}{Bike NYC} 
& \multicolumn{3}{c|}{Taxi BJ} 
& \multicolumn{3}{c}{Spain OD} \\
\multicolumn{2}{c|}{} 
& RMSE & CPC & MAE 
& RMSE & CPC & MAE 
& RMSE & CPC & MAE \\
\midrule

\multicolumn{11}{l}{\textit{Traditional time-series baselines}} \\
& MA        & 19.87 & 0.37 & 5.28  & 46.12 & 0.39 & 33.26 & 52.39 & 0.29 & 29.16 \\
& ARIMA     & 12.65 & 0.42 & 4.51  & 25.98 & 0.40 & 19.87 & 45.66 & 0.34 & 28.97 \\
& VAR       & \underline{12.50} & \underline{0.42} & \underline{4.48}  & \underline{25.61} & \underline{0.40} & \underline{19.33} & \underline{43.84} & \underline{0.36} & \underline{28.83} \\
\midrule

\multicolumn{11}{l}{\textit{Deep learning baselines}} \\
& RNN       & 10.24 & 0.58 & 4.15  & 20.04 & 0.45 & 16.31 & 30.06 & 0.43 & 18.20 \\
& LSTM      & 10.29 & 0.59 & 4.16  & 20.00 & 0.47 & 16.04 & 29.78 & 0.42 & 18.98 \\
& GRU       & 10.15 & 0.59 & 4.10  & 19.99 & 0.44 & 16.15 & 29.69 & 0.43 & 17.65 \\
& ST-ResNet & 9.38  & 0.61 & 4.22  & 19.32 & 0.48 & 15.26 & --     & --    & --     \\
& DMVSTNet  & 9.16  & 0.64 & 4.17  & 18.89 & 0.49 & 15.20 & --     & --    & --     \\
& ACMF      & 8.88  & 0.65 & 4.13  & 18.28 & 0.49 & 15.08 & --     & --    & --     \\
& CrowdNet  & 8.53  & 0.67 & 4.07  & 14.91 & 0.54 & 12.08 & 28.22 & \underline{0.45} & 14.26 \\
& DSTAGNN   & 8.15  & 0.67 & 3.68  & 14.35 & 0.56 & 12.00 & 28.10 & \underline{0.45} & 13.58 \\
& MSAGGN    & \underline{8.02}  & \underline{0.68} & \underline{3.59}  & \underline{14.12} & \underline{0.58} & \underline{11.95} & \underline{28.05} & \underline{0.45} & \underline{13.26} \\
\midrule

\multicolumn{11}{l}{\textit{Foundation models, zero-shot}} \\
& TimeLLM  & 6.52 & 0.69 & 3.16 & 9.82 & 0.61 & 8.22 & 24.18 & 0.59 & 10.42 \\
& Moirai-S & 6.32 & 0.70 & 3.08 & 9.58 & 0.61 & 7.98 & 23.68 & 0.62 & 10.12 \\
& TimesFM  & 6.18 & 0.70 & \underline{3.04} & 9.46 & \underline{0.62} & 7.88 & 23.29 & 0.61 & 9.97  \\
& Moirai-L & \underline{6.09} & \underline{0.72} & \underline{3.04} & \underline{9.32} & \underline{0.62} & \underline{7.34} & \underline{21.74} & \underline{0.67} & \underline{9.94}  \\
\midrule

\multicolumn{11}{l}{\textit{Foundation models, fine-tuned}} \\
& Moirai-S$_{\mathrm{FT}}$ & 6.28 & 0.72 & 3.00 & 9.46 & 0.63 & 7.75 & 22.08 & 0.64 & 10.01 \\
& TimesFM$_{\mathrm{FT}}$  & 6.11 & 0.71 & \underline{2.98} & 9.42 & \underline{0.64} & 7.71 & 21.72 & 0.64 & 9.86 \\
& Moirai-L$_{\mathrm{FT}}$ & \underline{6.01} & \underline{0.74} & 2.99 & \underline{9.11} & \underline{0.64} & \underline{7.26} & \underline{21.55} & \underline{0.67} & \underline{9.85} \\
\midrule

\multicolumn{2}{l|}{\textbf{TS-Mob}}
& \textbf{5.87} & \textbf{0.75} & \textbf{2.86}
& \textbf{8.95} & \textbf{0.67} & \textbf{7.08}
& \textbf{20.81} & \textbf{0.69} & \textbf{9.50} \\
\bottomrule
\end{tabular}%
}
\caption{Main forecasting results on Bike NYC, Taxi BJ, and Spain OD. RMSE and MAE are lower-is-better, while CPC is higher-is-better. Baselines are grouped into traditional time-series methods, deep learning models, zero-shot foundation models, and fine-tuned foundation models. Underlined values indicate the best baseline within each group and dataset/metric, while bold values indicate the best overall result. FT denotes fine-tuned. Missing entries for Spain OD correspond to grid-convolutional models that require regular spatial tessellations and are therefore not directly applicable to the irregular administrative tessellation of the Spain dataset.}
\label{tab:results}
\end{table*}

Table~\ref{tab:results} reports the performance of all baselines and TS-Mob on three datasets (Bike NYC, Taxi BJ, Spain OD) using RMSE and MAE (lower is better) and CPC (higher is better). A clear and consistent pattern emerges across all settings: classical statistical baselines (MA, ARIMA, VAR) provide a reasonable but limited reference point; specialized deep spatio-temporal models significantly improve upon them; foundation-model-based approaches (TimeLLM, Moirai, TimesFM) deliver another pronounced leap in accuracy; and our TS-Mob achieves the best results on every dataset and metric, establishing state-of-the-art performance. Critically, the gains obtained by TS-Mob are not confined to error magnitude (RMSE/MAE), but also translate into higher CPC, indicating better recovery of the OD structure rather than merely improved point-wise regression. This pattern is most pronounced on Spain OD, where the irregular tessellation excludes grid-based architectures entirely, the destination set spans 2,735 administrative tiles, and the complexity of the spatial structure makes explicit geographic conditioning most valuable. 

Starting from Bike NYC, the classical Moving Average (MA) baseline is clearly insufficient (RMSE $19.87$, CPC $0.37$, MAE $5.28$), while ARIMA and VAR substantially reduce error (RMSE $12.65$, CPC $0.42$). Among classical approaches, VAR is best (RMSE $12.50$, MAE $4.48$, CPC $0.42$), yet TS-Mob improves over it, reducing RMSE by up to a $53.04\%$ and MAE up to a $36.16\%$, and increasing CPC by $78.57\%$ (from $0.42$ to $0.75$). Deep spatio-temporal models further strengthen performance: ST-ResNet, DMVSTNet, and ACMF push CPC into the $0.61$ to $0.65$ range, and graph-based methods bring additional gains, with MSAGGN representing the strongest deep baseline on Bike NYC (RMSE $8.02$, CPC $0.68$, MAE $3.59$). TS-Mob still provides a substantial margin over this best deep competitor: RMSE drops by $26.81\%$, MAE by $20.33\%$, and CPC increases by $10.29\%$ (from $0.68$ to $0.75$). Concerning foundation models, TimeLLM (the only text-based baseline) already reaches RMSE $6.52$ and CPC $0.69$, while Moirai-S and TimesFM further improve, with Moirai-L as the best foundation baseline (RMSE $6.09$, CPC $0.72$, MAE $3.04$). Even in this high-performance regime, TS-Mob continues to yield consistent gains, improving over Moirai-L by $3.61\%$ in RMSE and $4.16\%$ in CPC (from $0.72$ to $0.75$), while also achieving the lowest MAE ($2.86$). This indicates that the explicit geographic/social attractiveness conditioning remains complementary to the temporal prior learned through large-scale pre-training.

Taxi BJ exhibits the same ordering across model families, with even sharper separations. The MA baseline performs poorly (RMSE $46.12$, CPC $0.39$, MAE $33.26$), and while ARIMA/VAR reduce error substantially (best classical: VAR, RMSE $25.61$, CPC $0.40$, MAE $19.33$), they remain far from competitive with deep and foundation approaches. Relative to VAR, TS-Mob reduces RMSE by $65.05\%$ and MAE by $63.37\%$, and increases CPC by $67.50\%$ (from $0.40$ to $0.67$), underscoring how strongly the taxi setting benefits from richer modeling. Deep architectures offer large improvements, especially graph-based ones: MSAGGN is the best deep baseline (RMSE $14.12$, CPC $0.58$, MAE $11.95$), yet TS-Mob still improves  over it (RMSE $-36.61\%$, MAE $-40.75\%$, CPC $+15.51\%$). Foundation-model approaches again form the strongest baseline group: TimeLLM reaches RMSE $9.82$ (CPC $0.61$), TimesFM reaches RMSE $9.46$ with CPC $0.62$, and Moirai-L improves RMSE to $9.32$ with MAE $7.34$ and CPC $0.62$. TS-Mob surpasses them all with RMSE $8.95$, CPC $0.67$, MAE $7.08$, corresponding to a $3.96\%$ RMSE reduction versus the best RMSE foundation baseline (Moirai-L), a $3.54\%$ MAE reduction, and a CPC improvement of $8.06\%$ over the best CPC foundation baseline. The consistency of these gains across metrics suggests that the attractiveness covariate is not merely acting as a regularizer, but is adding structured information that helps the temporal model better allocate flow intensity across destinations.

Spain OD is the most challenging setting due to its larger geographic scope, and it further highlights the importance of scalable priors and explicit structural conditioning. The Spain OD setting is particularly revealing: grid-based architectures are inapplicable to its irregular tessellation, yet TS-Mob achieves the largest relative gains over foundation baselines precisely where the structural conditioning is most needed. Classical methods lag substantially (best classical: VAR with RMSE $43.84$, CPC $0.36$, MAE $28.83$), and TS-Mob improves over VAR with a $52.53\%$ RMSE reduction, a $67.04\%$ MAE reduction, and a $91.66\%$ CPC increase (from $0.36$ to $0.69$). Among the deep graph-based models reported, CrowdNet, DSTAGNN, and MSAGGN form a tight cluster (up to RMSE $28.22$, CPC $0.45$), indicating that learning spatial dependencies helps but reaches a ceiling in recovering OD structure at this scale (CPC remains fixed at $0.45$ for all three). TS-Mob breaks this ceiling decisively, reducing RMSE by $25.81\%$ and MAE by $28.35\%$ relative to the best deep baseline (MSAGGN), and increasing CPC by $53.33\%$ (from $0.45$ to $0.69$). Foundation models again outperform deep baselines (TimeLLM RMSE $24.18$, Moirai-S $23.68$, TimesFM $23.29$, Moirai-L $21.74$), with Moirai-L being the best foundation baseline overall. TS-Mob achieves RMSE $20.81$, CPC $0.69$, MAE $9.50$, improving over Moirai-L by $4.27\%$ in RMSE and $4.42\%$ MAE and by $2.98\%$ in CPC. Compared to TimesFM, TS-Mob yields even larger relative gains on Spain OD, suggesting that explicitly providing a gravity-inspired attractiveness signal is particularly valuable when the spatial structure is complex and the destination set is large.

A group-wise comparison using only Table~\ref{tab:results} further clarifies the contribution of TS-Mob relative to each baseline family. Against the strongest classical baseline (VAR), TS-Mob reduces RMSE by $53.04\%$ (Bike), $65.05\%$ (Taxi), and $52.53\%$ (Spain), while increasing CPC by $78.57\%$, $67.50\%$, and $91.66\%$, respectively. Against the strongest deep baseline (MSAGGN where available), TS-Mob reduces RMSE by $26.81\%$ (Bike), $36.61\%$ (Taxi), and $28.35\%$ (Spain), while improving CPC by $10.29\%$, $15.51\%$, and $53.33\%$. Finally, even against the strongest foundation baseline (typically Moirai-L for RMSE/MAE and TimesFM for CPC in Taxi), TS-Mob yields smaller but robust gains: RMSE improves by $3.61\%$ (Bike), $3.96\%$ (Taxi), and $4.27\%$ (Spain), while CPC improves by $4.16\%$, $8.06\%$, and $2.98\%$.
Lastly, to enhance fairness, we further compare our model against fine-tuned foundation-model baselines. As expected, fine-tuning improves the foundation-model baselines compared with their zero-shot counterparts. However, TS-Mob remains consistently better than the fine-tuned baselines across all datasets and metrics.
On Bike NYC, the strongest fine-tuned baseline reaches RMSE $6.01$ and CPC $0.74$, while TS-Mob obtains RMSE $5.87$ and CPC $0.75$. On Taxi BJ, TS-Mob improves over the strongest fine-tuned baseline from RMSE $9.11$ to $8.95$ and from CPC $0.64$ to $0.67$. On Spain OD, TS-Mob improves from RMSE $21.55$ to $20.81$ and from CPC $0.67$ to $0.69$. The corresponding maximum relative gains over the strongest fine-tuned baseline are $3.43\%$ in RMSE, $4.69\%$ in CPC, and $4.35\%$ in MAE.
As expected, these gains are smaller than those observed against traditional or deep learning baselines, but they are the most relevant comparison for assessing the contribution of geographic conditioning. The results indicate that the destination-attractiveness signal does not replace temporal foundation-model fine-tuning; rather, it provides complementary OD-structural information that improves forecasting even when the temporal backbone is already strong.

These results support the central claim of the paper: large-scale temporal pre-training provides a strong backbone for mobility forecasting, but injecting explicit geographic and social structure through a gravity-inspired attractiveness covariate yields additional, consistent improvements across datasets, metrics, and spatial scales, leading to state-of-the-art forecasting performance.


\subsection{Temporal Pattern Analysis}
\label{sec:temporal_pattern_analysis}

\begin{table}[t]
\centering
\small
\setlength{\tabcolsep}{9pt}
\renewcommand{\arraystretch}{0.90
}
\begin{tabular}{lcccc}
\toprule
Dataset & Horizon & RMSE & CPC & MAE \\
\midrule
\multirow{6}{*}{Bike NYC}
& 1  & 5.87 & 0.75 & 2.86 \\
& 2  & 5.91 & 0.75 & 2.95 \\
& 3  & 5.94 & 0.75 & 3.01 \\
& 4  & 5.99 & 0.74 & 3.03 \\
& 5  & 6.04 & 0.74 & 3.08 \\
& 10 & 6.11 & 0.72 & 3.11 \\
\midrule
\multirow{6}{*}{Taxi BJ}
& 1  & 8.95 & 0.67 & 7.08 \\
& 2  & 9.11 & 0.65 & 7.15 \\
& 3  & 9.18 & 0.62 & 7.18 \\
& 4  & 9.30 & 0.61 & 7.20 \\
& 5  & 9.38 & 0.61 & 7.28 \\
& 10 & 9.47 & 0.59 & 7.33 \\
\midrule
\multirow{6}{*}{Spain OD}
& 1  & 20.81 & 0.69 & 9.50 \\
& 2  & 21.21 & 0.67 & 9.68 \\
& 3  & 21.55 & 0.66 & 9.92 \\
& 4  & 21.94 & 0.66 & 10.07 \\
& 5  & 22.30 & 0.64 & 10.16 \\
& 10 & 22.99 & 0.60 & 10.23 \\
\bottomrule
\end{tabular}
\caption{Multi-horizon forecasting results for TS-Mob. Lower RMSE/MAE and higher CPC are better.}
\label{tab:multihorizon}
\end{table}

\begin{figure*}[ht!]
    \centering
    \includegraphics[width=0.99\textwidth]{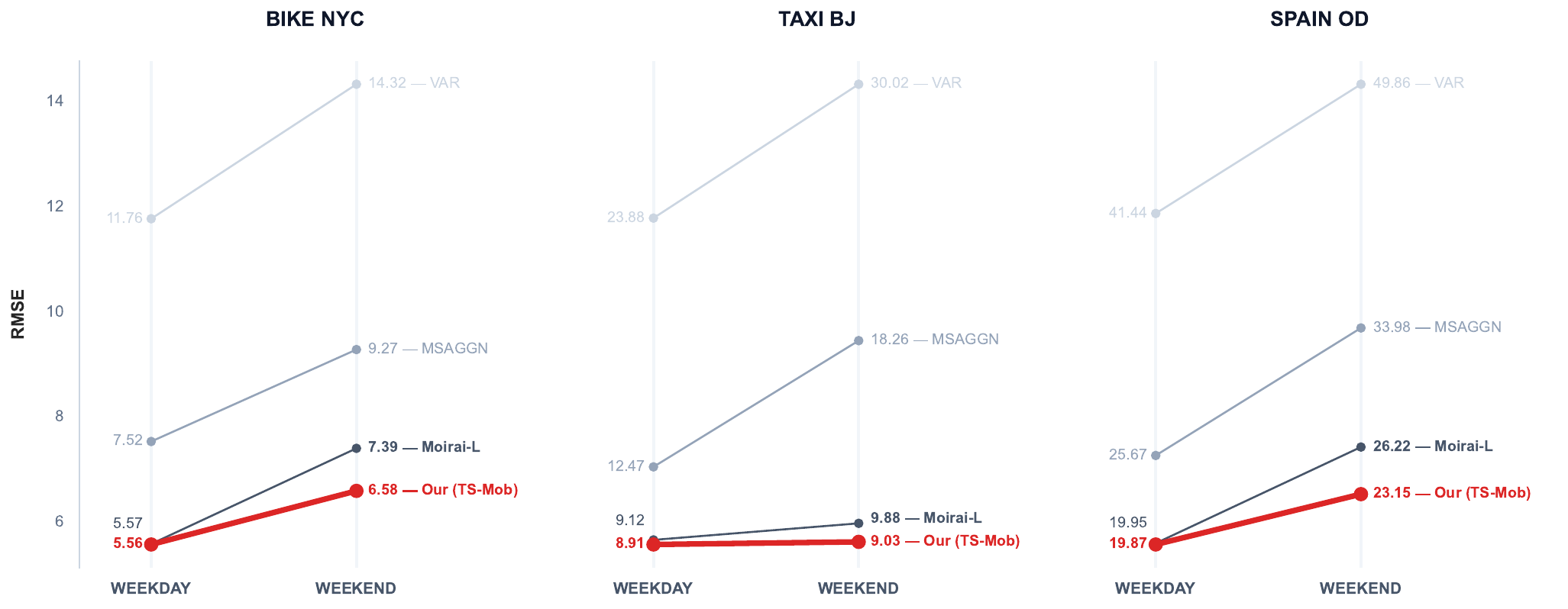}
    \caption{RMSE comparison of TS-Mob against the strongest representative from each baseline family by weekdays vs.\ weekends.}

    \label{fig:wewd}
\end{figure*}

We further analyze TS-Mob under two more demanding forecasting conditions: longer prediction horizons and weekday/weekend temporal regimes. 

Figure~\ref{fig:wewd} reports weekdays vs.\ weekends RMSE for TS-Mob and, for each baseline family in Table~\ref{tab:results}, the strongest representative method: VAR for classical forecasting, MSAGGN for deep spatio-temporal learning, and Moirai-L for time-series foundation models. Across all datasets and methods, weekends are consistently more challenging than weekdays, reflecting that weekend mobility is less dominated by regular commuting patterns and exhibits a broader and more heterogeneous destination mix.
On Bike NYC, TS-Mob increases from $5.56$ (weekday) to $6.58$ (weekend), while the corresponding degradation is larger for VAR ($11.76$ to $14.32$) and MSAGGN ($7.52$ to $9.27$); Moirai-L also degrades ($5.57$ to $7.39$). On Taxi BJ, TS-Mob remains stable across regimes ($8.91$ to $9.03$), and substantially outperforms VAR ($23.88$ to $30.02$) and MSAGGN ($12.47$ to $18.26$); Moirai-L shows a smaller weekend shift ($9.12$ to $9.88$). On Spain OD, TS-Mob increases from $19.87$ (weekday) to $23.15$ (weekend), compared to larger degradations for VAR ($41.44$ to $49.86$) and MSAGGN ($25.67$ to $33.98$); Moirai-L also degrades ($19.95$ to $26.22$). Overall, the stratified evaluation reinforces the main results: TS-Mob not only improves average accuracy, but also contains the weekend-induced performance drop more effectively than classical and deep baselines, and remains competitive with the strongest foundation-model baseline across datasets.

In addition, Table~\ref{tab:multihorizon} reports TS-Mob multi-horizon performance from one to ten steps ahead. The one-step setting remains the strongest, as expected, but performance degrades gradually rather than abruptly. On Bike NYC, RMSE increases from $5.87$ to $6.11$ and CPC decreases from $0.75$ to $0.72$ at ten steps. Taxi BJ shows a larger CPC decline, from $0.67$ to $0.59$, while RMSE increases from $8.95$ to $9.47$. Spain OD follows the same pattern, with RMSE increasing from $20.81$ to $22.99$ and CPC decreasing from $0.69$ to $0.60$. Overall, the results indicate that TS-Mob remains stable beyond the one-step setting, although longer horizons naturally make OD allocation harder.

\subsection{Ablation Study}
\label{sec:ablation}

We evaluate the contribution of the main TS-Mob components by comparing the full model with three variants. \textit{Mobility only} fine-tunes the TimesFM backbone using only historical OD flow values, without external covariates. \textit{Mobility + Weather} adds precipitation and temperature to the temporal forecaster, while excluding the geographic attractiveness signal. \textit{Mobility + Attractiveness} conditions the temporal forecaster on the gravity-inspired attractiveness signal, while excluding weather covariates. The full TS-Mob model combines historical mobility, weather, and attractiveness.

\begin{table}[t]
\centering
\scriptsize
\setlength{\tabcolsep}{3.2pt}
\renewcommand{\arraystretch}{1.5}
\resizebox{\columnwidth}{!}{%
\begin{tabular}{l|cc|cc|cc}
\toprule
& \multicolumn{2}{c|}{Bike NYC}
& \multicolumn{2}{c|}{Taxi BJ}
& \multicolumn{2}{c}{Spain OD} \\
Variant
& RMSE & CPC
& RMSE & CPC
& RMSE & CPC \\
\midrule
Mobility only
& 6.11 & 0.71
& 9.42 & 0.64
& 21.72 & 0.64 \\
+ Weather
& 6.05 & 0.74
& 9.28 & 0.65
& 21.33 & 0.67 \\
+ Attractiveness
& 5.91 & \textbf{0.75}
& 9.02 & 0.66
& 20.99 & \textbf{0.69} \\
Full TS-Mob
& \textbf{5.87} & \textbf{0.75}
& \textbf{8.95} & \textbf{0.67}
& \textbf{20.81} & \textbf{0.69} \\
\bottomrule
\end{tabular}%
}
\caption{Ablation study. RMSE is lower-is-better and CPC is higher-is-better. The mobility-only variant corresponds to fine-tuned TimesFM without exogenous covariates.}
\label{tab:ablation}
\end{table}

Table~\ref{tab:ablation} reports the ablation results in terms of RMSE and CPC, capturing both point-wise forecasting accuracy and the overlap between predicted and observed OD flows. The mobility-only variant already provides a strong reference point, since it corresponds to a fine-tuned TimesFM forecaster applied directly to OD flow histories. Adding weather improves RMSE across all datasets, from $6.11$ to $6.05$ on Bike NYC, from $9.42$ to $9.28$ on Taxi BJ, and from $21.72$ to $21.33$ on Spain OD. The corresponding CPC values also increase, especially on Bike NYC and Spain OD, indicating that short-term exogenous conditions help the temporal model adjust flow patterns.

The attractiveness signal produces a larger improvement than weather in most settings. Compared with the mobility-only variant, \textit{Mobility + Attractiveness} reduces RMSE from $6.11$ to $5.91$ on Bike NYC, from $9.42$ to $9.02$ on Taxi BJ, and from $21.72$ to $20.99$ on Spain OD. The effect is also visible in CPC: on Bike NYC, CPC increases from $0.71$ to $0.75$, matching the full model; on Spain OD, it increases from $0.64$ to $0.69$, again matching the full model. This pattern suggests that the attractiveness signal is especially useful for allocating predicted mobility volume across destinations, which is consistent with its role as an OD-specific geographic prior.

The full model achieves the best RMSE on all datasets and the best or tied-best CPC. Combining weather and attractiveness reduces RMSE relative to the mobility-only variant by $3.93\%$ on Bike NYC, $4.99\%$ on Taxi BJ, and $4.19\%$ on Spain OD. The comparison highlights the complementary role of the two sources of conditioning: weather captures short-term exogenous variation, while attractiveness captures geographic and activity-based destination plausibility.

\section{Discussion and Conclusions}
\label{sec:conclusion}

This work introduced TS-Mob, a mobility forecasting framework that couples a time series foundation model with an explicit and interpretable representation of geographic and social structure. The core idea is to provide the temporal forecaster with an origin-destination attractiveness signal computed through a gravity-inspired neural formulation from open data sources. This conditioning injects geographic and social priors—distance-based separation costs, population mass effects, and POI-derived proxies of opportunities—directly into the forecasting pipeline, allowing the temporal model to focus on short-term dynamics and exogenous drivers such as weather.

Across three real-world datasets spanning different spatial scales and temporal resolutions (Bike NYC, Taxi BJ, and nation-scale Spain OD), TS-Mob achieves state-of-the-art performance on short-term crowd-flow prediction, outperforming classical forecasting baselines, widely adopted deep spatio-temporal architectures, and strong foundation-model baselines (TimeLLM, Moirai, TimesFM). The results highlight three main empirical takeaways. First, time series foundation models provide a strong temporal backbone for mobility: even when used without explicit spatial inductive biases, they consistently surpass classical and deep baselines. Second, explicitly conditioning the temporal block on gravity-inspired attractiveness yields additional and consistent gains across datasets and metrics, including CPC improvements that indicate better recovery of OD structure rather than only reduced point-wise error. Third, TS-Mob remains robust across temporal regimes: the stratified analysis shows that performance degrades as expected on weekends, yet TS-Mob maintains the best or near-best performance and contains the degradation more effectively than classical and deep competitors.


The study also suggests several directions for future work. First, while TS-Mob already leverages weather and a temporally varying population proxy through the attractiveness mechanism, additional dynamic context could further improve robustness under disruptions, such as major events, strikes, school calendars, and road interventions. These signals could be integrated either as additional covariates to the temporal block or as inputs to the attractiveness estimator when they directly affect destination appeal. Second, our current conditioning uses a compact OD-specific attractiveness stream; richer conditioning could expose multiple interpretable channels or use learned low-dimensional embeddings that preserve interpretability while improving capacity. Third, the gravity-inspired block is trained first and then used to provide covariates for fine-tuning the temporal model. Exploring joint or alternating optimization may yield further gains, but it requires care to avoid inadvertently encoding temporal information into the structural component. 

Finally, broader evaluation would strengthen the evidence of portability. While we cover three datasets with diverse characteristics and include strong baselines, future work should test additional cities, transport modes, and tessellations, and consider complementary evaluation protocols such as ranking-based destination allocation metrics (top-$k$ destinations), tail-focused errors for extreme-demand periods, and computational efficiency in production-like settings.

In conclusion, TS-Mob demonstrates that conditioning a time series foundation model on an explicit gravity-inspired attractiveness signal computed from open data yields a robust and interpretable approach to short-term origin-destination forecasting. The results support a practical path toward mobility predictors that are less dependent on dense historical mobility data, more portable across geographic contexts, and easier to inspect and reason about, while achieving state-of-the-art performance across different datasets and temporal regimes.


\clearpage
\bibliographystyle{unsrtnat}
\bibliography{sample-base}

\clearpage
\appendix

\section{Overture POI Top-Level Categories and Data Mapping}
\label{app:overture_categories}

Data for Points of Interest (POIs) are obtained through the DuckDB interface available on the Overture Maps website. Each element downloaded, in addition to other information, has a latitude, a longitude, and a high-level category. Latitude and longitude are used to map each POI into a region of interest. After, for each region of interest and for each top-level category, we count the number of POIs to obtain their density.   
Overall, Overture data comes with 16 top-level categories:
\vspace{2pt}

\begin{center}
\small
\begin{tabular}{@{}ll@{}}
\texttt{accommodation} & \texttt{financial\_service}\\
\texttt{active\_life} & \texttt{health\_and\_medical}\\
\texttt{arts\_and\_entertainment} & \texttt{professional\_services}\\
\texttt{attractions\_and\_activities} & \texttt{public\_service\_and\_government}\\
\texttt{automotive} & \texttt{real\_estate}\\
\texttt{beauty\_and\_spa} & \texttt{religious\_organization}\\
\texttt{eat\_and\_drink} & \texttt{retail}\\
\texttt{education} & \texttt{travel}\\
\end{tabular}
\end{center}

\FloatBarrier

\section{Reproducibility Details}
\label{app:repro_details}
In this Section, we report all the necessary details to reproduce the experiments carried out in the paper. 

\subsection{Data Preprocessing}
\label{app:data_prepro}
In our paper, we use three different types of data sources. 

\subsubsection{Points of Interest}
Concerning POIs, as we already discussed in Appendix A, data are retrieved through the DuckDB interface available on Overture's website. Each POI is associated with a category $c$, a latitude, and a longitude. The category $c$ is among the 16 categories listed in Appendix A. 
For each location $i$ and for each category $c$, we count the number of POIs forming, for each location $i$, a vector $\vec{c_i}$ of size 16, where each element correspond to the count of POIs for a specific category in location $i$. 

\subsubsection{Weather Data}
Historical weather information related to perceived temperature and precipitation levels in millimeters is obtained via the Open Meteo APIs \footnote{https://open-meteo.com/}. Importantly, data are provided with hourly aggregations. While mobility data in Spain OD and Bike New York are also hourly aggregated, the data collected from taxis in Beijing has a temporal aggregation of 30 minutes. Thus, during fine-tuning and inference time, weather information is repeated two times and updated every two steps.

\subsubsection{Mobility Data}
Taxi Beijing and Bike New York City are processed and prepared exactly as described in \cite{zhang2017deep}. We report a description of the procedure here but further details are available in the paper. 

To enable the modeling of citywide crowd dynamics using convolutional architectures, we first transform raw trajectory data into a structured spatio-temporal format. The geographic area of interest is partitioned into an $I \times J$ grid based on longitude and latitude coordinates, where each cell $(i, j)$ represents a specific functional region of the city.

For each grid cell at a discrete time interval $t$, we define two primary types of crowd flows based on the movement of objects (e.g., vehicles, pedestrians):(i) inflow is the total volume of traffic entering cell $(i, j)$ from any other location during interval $t$. (ii) Outflow is the total volume of traffic leaving cell $(i, j)$ to any other destination during interval $t$.

Mathematically, the state of the city at time $t$ is represented as a tensor $\mathbf{X}_t \in \mathbb{R}^{2 \times I \times J}$. The two channels of this tensor capture the inflow and outflow intensity across the spatial grid, effectively treating the city's mobility state as a two-channel image. This transformation allows the model to leverage convolutional layers to capture local spatial correlations and the dependencies between neighboring regions. For the Beijing dataset (TaxiBJ), the city is mapped to a $32 \times 32$ grid, while the New York City dataset (BikeNYC) utilizes a $16 \times 8$ grid configuration.

The same procedure is followed for the Spanish data, which, however, are already divided into $N$ tiles with $N=2735$. All the tiles have irregular shapes, and thus, as mentioned in the main text, deep learning baselines based on convolutions will not be used with this dataset. In addition, the dataset comes already in an aggregated form: for each pair of origin-destination $i,j$, the number of people traveling from $i$ to $j$ is provided.

\subsection{Baseline Training}
\label{app:baselines}

All the selected baselines are available through official repositories or libraries. Whenever default hyperparameters are used is because models are already trained and tested on Bike NYC and taxi Beijing: 
\begin{itemize}
    \item MA, ARIMA, VAR are available in statsmodels
    \item ST-ResNet is available at https://github.com/topazape/ST-ResNet and the default parameters have been used
    \item DMVSTNet is available at \\https://github.com/huaxiuyao/DMVST-Net \\ and the hyperparameters of ST-ResNet have been used for the training
    \item ACMF code is at \\https://github.com/liulingbo918/ATFM/tree/master and default parameters have been adopted.
    \item CrowdNet was trained using the default hyperparameters and the code is available at \\ https://github.com/jonpappalord/crowd\_flow\_prediction
    \item DSTAGNN is available at \\https://github.com/SYLan2019/DSTAGNN  \\ and default parameters have been used.
\end{itemize}
Foundation models have been used in a zero-shot way
\begin{itemize}
    \item TimesFM is available at \\ https://github.com/google-research/timesfm
    \item Moirai can be found at https://huggingface.co/Salesforce/moirai-1.0-R-large
    \item TimeLLM is at https://github.com/KimMeen/Time-LLM
\end{itemize}

Adjusted learning rates for Spain OD are listed in Table \ref{tab:hyper-spain}. Concerning the training and test split for the dataset, we followed the procedure indicated in \cite{zhang2017deep}. Thus, to evaluate the predictive performance of \textit{all} the models, including TS-Mob, we employ a chronological data splitting strategy. 
For our experimental validation, we partition the datasets as follows:

\begin{itemize}
    \item \textbf{TaxiBJ (Beijing):} The dataset is split such that the final 20\% of records are reserved for the testing set. All preceding months of data are utilized for model training and parameter optimization.
    \item \textbf{BikeNYC (New York):} Given the smaller temporal footprint of this dataset, the final 10 days of available records are selected as the testing set, while the remaining prior data serves as the training set.
    \item \textbf{Spain OD:} we kept the last 3 weeks for testing and the first 2 months and 1 week for training. 
\end{itemize}

During the training phase, we further segment the training data by reserving a small portion (e.g., the most recent 10\% points) as a validation set. This subset is used for hyperparameter tuning and early stopping to mitigate overfitting. The final evaluation is conducted on the held-out test set to report the model's generalized performance.

\begin{table}[h]
\centering
\begin{tabular}{lc}
\hline
\textbf{Model} & \textbf{Learning Rate} \\ \hline
ST-ResNet      & $1 \times 10^{-4}$     \\
DMVSTNet       & $1 \times 10^{-4}$     \\
ACMF           & $5 \times 10^{-4}$     \\
CrowdNet       & $1 \times 10^{-3}$     \\
DSTAGNN        & $1 \times 10^{-4}$     \\
TimesFM        & N/A        \\
Moirai         & N/A        \\
TimeLLM        & N/A    \\ \hline
\end{tabular}
\caption{Learning rates for Spain OD.}
\label{tab:hyper-spain}
\end{table}

\subsection{TimesFM Finetuning}
\label{app:times_finetuned}
In Table \ref{tab:time-fine} we summarize the settings for the finetuning of TimesFM.

\begin{table}[h]
\centering
\begin{tabular}{ll}
\hline
\textbf{Setting Category} & \textbf{Value} \\ \hline
Optimizer                 & SGD \\
Linear Warmup Epochs      & 25 \\
Total Epochs              & 50 \\
Learning Rate             & $5 \times 10^{-4}$ \\
Momentum                  & 0.9 \\
Gradient Clip             & 1.0 \\
Input Length              & 32 \\
\end{tabular}
\caption{Finetuning hyperparameters for TimesFM.}
\label{tab:time-fine}
\end{table}

\subsection{Computational Setting}
\label{app:Computational_settings}
All the models and datasets have been processed on a server with 126 GB of RAM and 2 NVIDIA GeForce RTX 2080 Ti with 12 GB of memory each. 

\subsection{Classical Spatial Interaction Baselines}
\label{app:spatial_interaction_baselines}

Table~\ref{tab:gravity_radiation_appendix} reports Gravity and Radiation as
classical spatial-interaction reference models. Unlike the temporal forecasting
baselines in Table~\ref{tab:results}, these models estimate OD interaction
strength from spatial and aggregate regional attributes and do not explicitly
model temporal sequence dynamics. We therefore report them separately in the
appendix. Their lower performance suggests that static spatial-interaction
structure alone is insufficient for accurate time-varying OD flow forecasting.

\begin{table}[t]
\centering
\small
\setlength{\tabcolsep}{3.5pt}
\renewcommand{\arraystretch}{1.1}
\resizebox{\columnwidth}{!}{%
\begin{tabular}{l|ccc|ccc|ccc}
\toprule
& \multicolumn{3}{c|}{Bike NYC}
& \multicolumn{3}{c|}{Taxi BJ}
& \multicolumn{3}{c}{Spain OD} \\
Model
& RMSE & CPC & MAE
& RMSE & CPC & MAE
& RMSE & CPC & MAE \\
\midrule
Gravity   
& \textbf{11.28} & \textbf{0.59} & \textbf{4.42}
& \textbf{24.85} & \textbf{0.45} & \textbf{16.77}
& \textbf{40.41} & \textbf{0.42} & \textbf{25.79} \\
Radiation 
& 12.02 & 0.51 & 4.49
& 24.99 & 0.39 & 18.92
& 41.03 & 0.35 & 27.82 \\
\bottomrule
\end{tabular}%
}
\caption{Classical spatial interaction baselines. Gravity and Radiation are reported as spatial-interaction reference models rather than direct temporal forecasting baselines.}
\label{tab:gravity_radiation_appendix}
\end{table}

\end{document}